\documentclass[conference]{IEEEtran}
\IEEEoverridecommandlockouts
\usepackage{cite}
\usepackage{amsmath,amssymb,amsfonts}
\usepackage{graphicx}
\usepackage{textcomp}
\usepackage{xcolor}
\usepackage{multicol}
\usepackage{color}
\usepackage{xspace}
\usepackage{enumerate}
\usepackage{mathtools}
\usepackage{graphicx}
\usepackage{amsmath}
\usepackage{amssymb}
\usepackage{booktabs}
\usepackage{xcolor}
\usepackage[ruled]{algorithm2e}
\usepackage{algpseudocode}
\usepackage{multirow}
\usepackage{epsfig}
\usepackage{subfigure} 
\usepackage{array}
\usepackage{dsfont}
\usepackage{tabularx}
\usepackage{threeparttable}
\usepackage{ragged2e}
\usepackage{colortbl}
\usepackage{changes}
\usepackage{eucal}
\usepackage{url}
\usepackage{caption}
\usepackage{xcolor}

\def\BibTeX{{\rm B\kern-.05em{\sc i\kern-.025em b}\kern-.08em
    T\kern-.1667em\lower.7ex\hbox{E}\kern-.125emX}}
\begin{document}

\title{Robust 3D Face Alignment with \\ Multi-Path Neural Architecture Search
\thanks{This work is supported by the National Natural Science Foundation of China (62302093), Jiangsu Province Natural Science Fund (BK20230833) and the Southeast University Start-Up Grant for New Faculty (RF1028623063). This work is also supported by the Big Data Computing Center of Southeast University. \\ \textsuperscript{*} Corresponding Author.}
}

\author{
 \IEEEauthorblockN{1\textsuperscript{st} Zhichao Jiang}
\IEEEauthorblockA{\textit{Institute of Deep Learning (IDL)} \\
\textit{Baidu}\\
Beijing, China \\
chao\_jichao@hotmail.com} 
\and
\IEEEauthorblockN{2\textsuperscript{nd} Hongsong Wang\textsuperscript{*}} 
\IEEEauthorblockA{\textit{Department of Computer Science and Engineering} \\
	\textit{Key Laboratory of New Generation Artificial Intelligence} \\
	\textit{Technology and Its Interdisciplinary Applications} \\
Southeast University, Nanjing, China \\
hongsongwang@seu.edu.cn} 
\and
\hspace{50pt}\IEEEauthorblockN{3\textsuperscript{rd} Xi Teng}
\IEEEauthorblockA{\hspace{50pt}\textit{Computer Vision Technology Institution} \\
\hspace{50pt}\textit{Baidu}\\
\hspace{50pt} Beijing, China \\
\hspace{50pt} xiteng01@baidu.com} 
\and
\hspace{40pt}\IEEEauthorblockN{4\textsuperscript{th} Baopu Li} 
\IEEEauthorblockA{\hspace{40pt}\textit{Baidu Research} \\
\hspace{40pt}\textit{Baidu}\\
\hspace{40pt}Sunnyvale, USA \\
\hspace{40pt}bpli.cuhk@gmail.com} 
}

\maketitle

\begin{abstract}
	3D face alignment is a very challenging and fundamental problem in computer vision. Existing deep learning-based methods manually design different networks to regress either parameters of a 3D face model or 3D positions of face vertices. However, designing such networks relies on expert knowledge, and these methods often struggle to produce consistent results across various face poses. To address this limitation, we employ Neural Architecture Search (NAS) to automatically discover the optimal architecture for 3D face alignment. We propose a novel Multi-path One-shot Neural Architecture Search (MONAS) framework that leverages multi-scale features and contextual information to enhance face alignment across various poses. The MONAS comprises two key algorithms: Multi-path Networks Unbiased Sampling Based Training and Simulated Annealing based Multi-path One-shot Search. Experimental results on three popular benchmarks demonstrate the superior performance of the MONAS for both sparse alignment and dense alignment.
\end{abstract}


\section{Introduction}
\label{sec:intro}
3D face alignment~\cite{zhu2016face,Booth17}, which aims to estimate the precise 3D shape and pose of a face from a single 2D image, is essential for many face related problems. Compared to traditional 2D face alignment, this task offers a more precise and detailed representation of human faces.

Recently, data-driven approaches that utilize deep neural networks for training have garnered the interest of numerous researchers. Many methods, such as \cite{tran2018nonlinear,yuxiang2019,Tran2019TowardsHN}, employ convolutional neural networks (CNNs) to estimate the parameters of a parametric 3D face model, specifically 3DMM \cite{blanz1999morphable}.
The 3D model parameter regression based approaches such as~\cite{Tran2019TowardsHN} achieve high-fidelity face shape and dense face alignment in controlled settings. 
However, these approaches lack good scalability when it comes to dealing with challenges like occlusions, pose variations, and extreme lighting conditions in the wild, and their results heavily depend on the performance of the off-the-shelf 3D face model.

Another line of research involves directly regressing the 3D position of face vertices using UV map and UV position (\cite{feng2018joint}) or 3D volume (\cite{jackson2017large}). The PRNet~\cite{feng2018joint} regresses the complete 3D face structure along with semantic meaning of keypoints from a monocular image using the UV position map. However, significant pose variations still present a challenge in modeling the regression process from facial appearances. In addition, manually designed networks such as PRNet~\cite{feng2018joint} may not constitute an optimal combination of features that can effectively handle various poses simultaneously.

This paper aims to explore 3D face alignment using Neural Architecture Search (NAS). Given that features at different scales are sensitive to local facial regions and pose variations, we propose the first multi-path one-shot NAS method to automatically search for the optimal network architectures that fully leverage multi-scale features. Specifically, we introduce a novel multi-path search space that considers both network topology and convolution operation type, aiming to establish meaningful connections between features of different scales. We then propose the Multi-path Networks Unbiased Sampling (MNUS) approach to sample child networks. Based on MNUS, we design an efficient multi-path supernet training algorithm to thoroughly explore various combinations of multi-scale features. Lastly, we devise a Simulated Annealing based Multi-path One-shot Search (SAMOS) algorithm to efficiently identify the optimal child network architecture.

Our main contributions can be summarized as follows.
\begin{itemize}
\item We present the first method that explores NAS for 3D face alignment, which searches the optimal network that directly regresses 3D positions of face vertices.
\item We introduce a novel Multi-path One-shot Neural Architecture Search (MONAS) framework to fully leverage multi-scale features for representation learning.
\item To efficiently train the supernet and search the optimal network, we propose a novel Multi-path Networks Unbiased Sampling Based Training together with a Simulated Annealing based Multi-path One-shot Search (SAMOS) method.
\end{itemize}

\section{Related works}

\noindent\textbf{3D Face Alignment.}
Face alignment is the process of pinpointing key facial features in an image, which can be categorized into sparse face alignment and dense face alignment depending on the type of facial features. 
To align faces with significant pose variations, numerous 3D face fitting methodologies have been proposed in the literature~\cite{jourabloo2016large,zhu2016face}. These methods fit a 3D morphable model (3DMM)~\cite{blanz2003face} to a 2D facial image, allowing for accurate alignment even in cases of large pose variations.
Deng et al.~\cite{deng2019accurate} formulate an end-to-end training framework to regress the 3DMM parameters including shape and texture.
Zhu et al. \cite{zhu2016face} fit a dense 3D face model to the image using CNN, and synthesize training samples in profile views to address the issue of data labeling. Guo et al. \cite{guo2020towards} further extend this method with meta-joint optimization, achieving better performance and stability.
To obtain the geometric parameters, Koizumi  et al.~\cite{koizumi2020look} present an unsupervised training method which regresses linear coefficients instead of coordinates.
Recently, Ruan et al.~\cite{Ruan2021SADRNetSD} propose a self-aligned dual faces regression framework to deal with face pose variation and occlusions. Lei et al.~\cite{Lei2023AHR} present a hierarchical representation network for accurate and detailed face reconstruction. 
Li et al.~\cite{Li2023DSFNetDS} present a dual space fusion network for robust and unconstrained 3D face alignment.


\noindent\textbf{One-Shot Neural Architecture Search.} 
Neural Architecture Search (NAS) aims to automatically search for the optimal architecture of a neural network. To improve the search efficiency, one-shot NAS~\cite{brock2017smash,bender2018understanding} only train one super-network and all sub-networks can be rapidly evaluated through weight-sharing.
Differentiable architecture search~\cite{liu2018darts} and path sampling-based approaches~\cite{guo2019single} are two prominent frameworks for super-network training. The former alternately optimizes architecture parameters and network weights, while the latter prioritizes ensuring the fairness and robustness of super-network training.
However, most one-shot NAS approaches (e.g., ~\cite{bender2018understanding,guo2019single,yu2020bignas,cai2020once}) are specifically tailored to single-path neural architectures for relatively simple tasks. As far as we know, there is no existing NAS-based framework for 3D face alignment. In this paper, we present a groundbreaking exploration of one-shot NAS for 3D face alignment.


\section{Methodology}
\begin{figure}[t]
\centering
\includegraphics[width=1.0\linewidth]{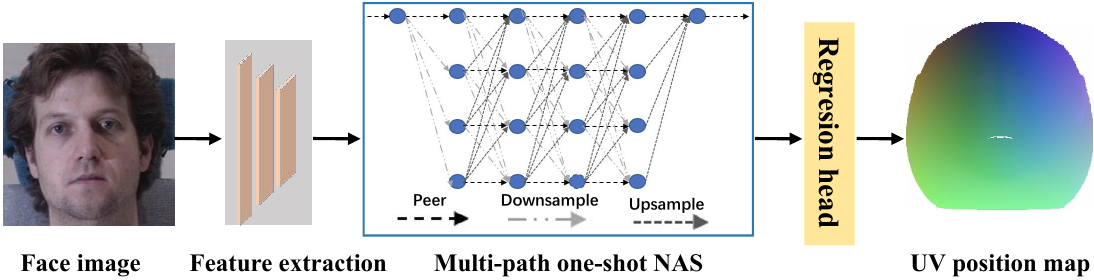}
\caption{The proposed framework of the multi-path neural architecture search for 3D face alignment. }
\label{fig:framework} %
\end{figure}

We utilize Neural Architecture Search (NAS) to find the optimal network structure for regressing the UV position map. A simple illustration of the proposed method is shown in Fig.~\ref{fig:framework}. The primary module is the Multi-path One-shot Neural Architecture Search (MONAS) which are described as follows.
\begin{figure}[t]
\centering
\includegraphics[width=1\linewidth]{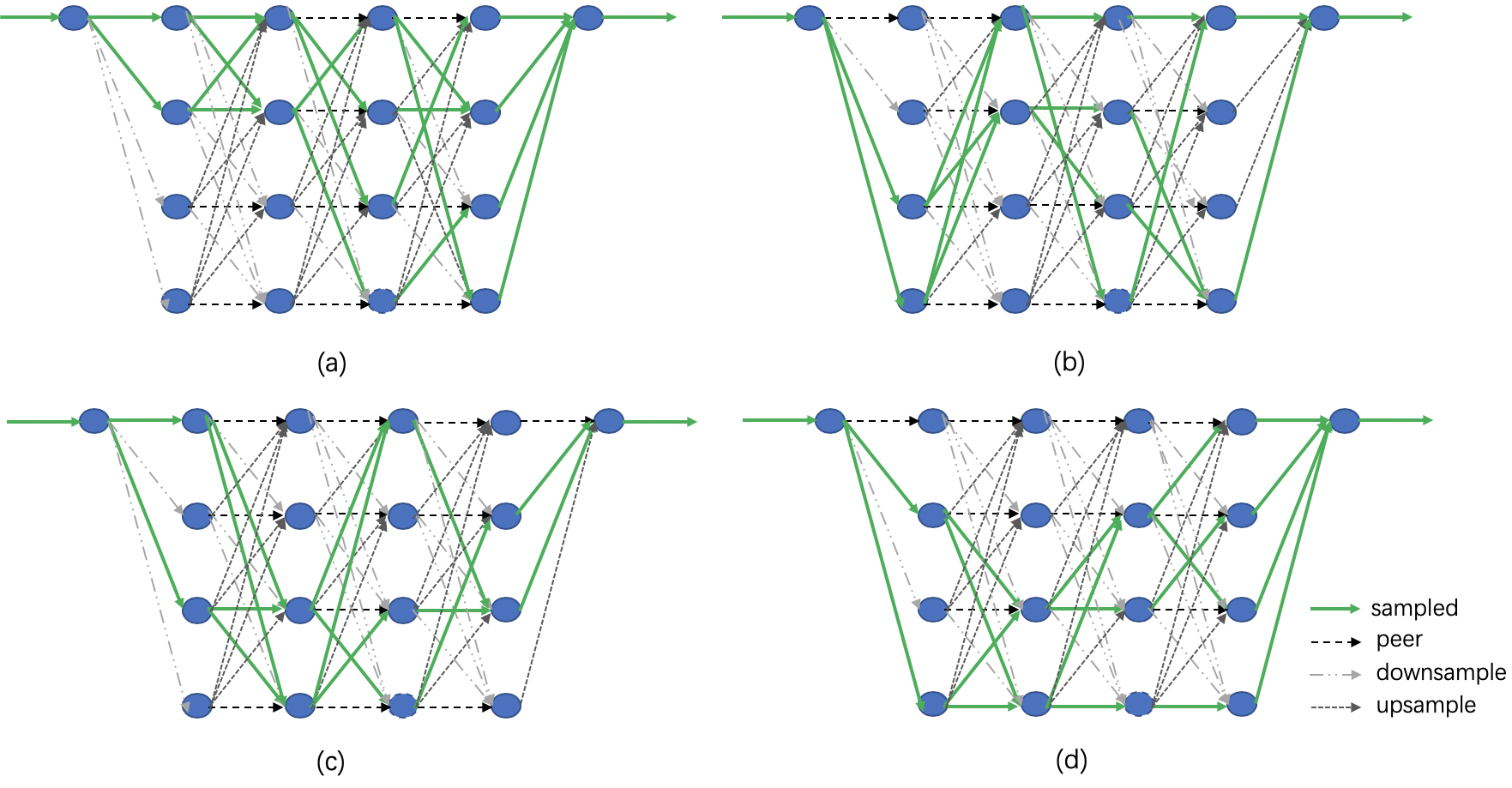}
\caption{A simplified example of the multi-path sampling, where (a)-(d) represent four sampled child networks. Nodes of blue circles in the same column belong to the same layer. Taking the third layer as an example, there are 16 unique connections in total between the third layer and the fourth layer, and each network has exactly four connections. Similarly, all connections are sampled exactly twice in the first layer. } 
\label{fig:sample1}
\end{figure}

\subsection{Network Search Space} \label{sec:searchspace}
To identify informative representations that work well for different poses and local regions, we carefully combine features in various scales. While existing multi-scale architectures have considered the fusion of features at adjacent scales, they are unable to cope with such variations effectively. Therefore, we propose a multi-path supernet that considers possible connections between features at different scales.
In our multi-path supernet-based search space, we consider both network topology and convolution operation type. 

The network topology is denoted by ${\mathcal T}$,
\begin{equation}
\mathcal T = [ c_{l,d_{out}}^{d_{in}}| l \in \mathcal{L}, d_{out},d_{in}   \in \mathcal{D}],
\label{eq:connection}
\end{equation}
where $\mathcal{L}=\{1,2,\cdots,L-1\}$, $\mathcal{D}=\{1,2,\cdots,D\}$, $L$ and $D$ are the numbers of layers and scales, respectively, and $c_{l,d_{out}}^{d_{in}}$ indicates the connection status between the $d_{in}$-th and $d_{out}$-th nodes in the $l$-th layer. Mathematically,
\begin{equation}
c_{l,d_{out}}^{d_{in}} \in \{0,1\}, \forall l \in \mathcal{L}, \forall d_{out},d_{in}   \in \mathcal{D}.
\label{eq:sub_connection}
\end{equation}

A node represents a summation operation of features at a specific scale. Within the same layer, if two nodes are adjacent and belong to the same column as depicted in Figure ~\ref{fig:framework}, the node located above has a scale twice as large as the one below it. Each node can be connected to all the nodes of different scales, and its feature map size can undergo downsampling, upsampling, or remain unchanged.

Note that we not only focus on what features should be fused, but also on how they should be fused. Therefore, we search for the best convolution for each connection. The convolution operations ${\mathcal C}$ is defined as
\begin{equation}
\mathcal C = [ b_{l,d_{out}}^{d_{in}}| l \in \mathcal{L}, d_{out},d_{in}   \in \mathcal{D}],
\label{eq:block}
\end{equation}
where the convolution type is
\begin{equation}
b_{l,d_{out}}^{d_{in}} \in \{HPM, 
identity\}, \forall l \in \mathcal{L}, \forall d_{out},d_{in}   \in \mathcal{D},
\label{eq:sub_block}
\end{equation}
where HPM is the convolution block type (hierarchical, parallel and multi-scale block) and $identity$ is the identity connection. We adopt the HPM block since we find that its performance is better than other common block types like basic block, residual block, depthwise block when we replace them with HPM block in any position.

The size of the search space without validity constraint of connections  is {$(|c|\times|t|-|c|+1)^{(L-2)\times{{|\mathcal{D}|}^{2}} + {|\mathcal{D}|}\times{2}}, c\in \mathcal{C}, t\in \mathcal{T}$}. 
We do not search for the kernel size due to the excessively large search space, and because a skillful combination of various scales has the potential to generate an equivalent effective receptive field, which is pivotal in addressing the challenge of face alignment.
It is worth mentioning that Hourglass \cite{newell2016stacked}, PRNet \cite{feng2018joint}, and HRNet \cite{sun2019deep} can be considered as special and manually designed cases of the proposed supernet.

\begin{algorithm}[t]
	\footnotesize
	\caption{Multi-path Networks Unbiased Sampling (MNUS) Based Training}
	\label{alg:alg1}
	\DontPrintSemicolon
	\KwIn{The initial multi-path supernet.\\}
	\KwOut{The well-trained multi-path supernet.\\}
	\label{alg:AEA}
	\DontPrintSemicolon
	\LinesNumbered
	\While{supernet not converge}
	{
		Sample $M^k(1), k=1,2,\cdots,K$ \\
		\For {$l=2,3,\cdots,L$} {
			Sample $M^k(l)$ based on prior layer's sampling.\\
		}
		\For {$k=1,2,\cdots,K$}{
			Accumulate the gradient of the supernet by $\mathcal M^k$. \\ 
		}
		Update weights of the supernet by accumulated gradients. \\
	}
\end{algorithm}

\subsection{Multi-Path Network Training}
For pose-invariant face alignment, our goal is to thoroughly explore various combinations of features at different scales. Therefore, it is crucial to ensure that all paths of feature fusion are given equal consideration. To address this issue, we propose a novel Multi-path Networks Unbiased Sampling (MNUS) based Training algorithm.  



The proposed MNUS process is outlined in Algorithm~\ref{alg:alg1}. During each iteration, we first sample $K$ child networks, and subsequently accumulate gradients of these child networks. Finally, parameters of the supernet are updated. 

Let $\mathcal M^k$ be the sampled child network, where $k \in \{1,2,\cdots,K\}$. We further analyze $\mathcal M^k$ by examining its individual layers
\begin{equation}
\mathcal M^k= [ M^k(l)| l \in \mathcal{L}], \forall k \in  \{1,2,\cdots,K\},
\label{eq:sample_divide}
\end{equation}
where $M^k(l)$ for the $l$-th layer is formulated as
\begin{equation}
M^k(l) = [ m_{l,d_{out}}^{k,d_{in}}| d_{out},d_{in}   \in \mathcal{D}], 
\label{eq:sub_sample_divide}
\end{equation}
where $m_{l,d_{out}}^{k,d_{in}}$ is composed of the connection status $c_{l,d_{out}}^{d_{in}}$ and the convolution type $b_{l,d_{out}}^{d_{in}}$ of the $k$-th child network.

The MNUS algorithm enforces that each connection between layer $l$ and layer $l+1$ are sampled exactly the same number of times by the $K$ child networks. Mathematically, $\forall \mathit{out1},\mathit{out2}, \mathit{in1},\mathit{in2} \in \mathcal{D}, \mathit{in1} \neq \mathit{in2}, \forall l \in \mathcal{L}$, the connections sampling process satisfies the following constraints,
\begin{equation}
\sum_{k=1}^{K}m_{l,d_{\mathit{out1}}}^{k,d_{\mathit{in1}}}=\sum_{k=1}^{K}m_{l,d_{\mathit{out2}}}^{k,d_{\mathit{in2}}},
\label{eq:connection fairness}
\end{equation}

The child network is subsequently sampled from the first layer to the last layer. Each layer has its preceding layer except the first layer. 
Thus, $\forall l > 1, \forall in,out \in \mathcal{D}$, we have:
\begin{align}
m_{l,d_{out}}^{k,d_{in}} <= max(m_{l-1,1}^{k,d_{out}},m_{l-1,2}^{k,d_{out}},\cdots, m_{l-1,D}^{k,d_{out}}), 
\label{eq:connection validness}
\end{align}
Eq. \ref{eq:connection validness} restricts that each node of a particular layer must have at least one connection from its previous layer. 
Examples of the proposed sampling strategy are shown in Fig.~\ref{fig:sample1}. In this way, connections belonging to the same layer are sampled exactly the same number of times by all child networks.

\begin{algorithm}[t!]
\footnotesize
\caption{Simulated Annealing based Multi-path One-shot Search}
\label{alg:SAMOS}
\DontPrintSemicolon
\KwIn{The well-trained multi-path supernet by MNUS.\\}
\KwOut{The best child network $\mathcal S$ and its corresponding penalty $p$.}
Initialize the child network $\mathcal S$ and calculate its penalty $p$, set the temperature $T^{\circ}$, the factor $\xi$ and the iteration number $k$.  \\
\While{ $k$ $>$ 0}
{
Generate neighbor child network $\mathcal S^k$ of $\mathcal S$ based on Eq. (7 - 8). \\
Evaluate the performance of the $\mathcal S^k$ and get the corresponding penalty $p_k$ based  on  the trained multi-path supernet. \\
Calculate $\Delta p \leftarrow p_k - p$.\\
\eIf { $ \Delta p < 0 $}
{
	Accept $\mathcal S$ as the feasible solution and set $p \leftarrow p_k$, $\mathcal S$ = $\mathcal S^k$.\\
}
{
	\If { $rand(0,1)<exp(-\Delta p/T^{\circ})) $}
	{
		Accept the neighbour $\mathcal S^k$ and set $p \leftarrow p_k$, $\mathcal S \leftarrow \mathcal S^k$.\\
	}
}
Update annealing temperature $T^{\circ} \leftarrow \xi \cdot T^{\circ}$. \\
Upate the iteration number $k \leftarrow k - 1$. \\
}
\end{algorithm}

\subsection{Optimal Network Search}
After the supernet is well-trained, identifying the optimal child network architecture remains a challenge due to the large number of candidate sub-networks in the supernet. To strike a balance between search efficiency and closeness to the optimal solution, we propose the Simulated Annealing based Multi-path One-shot Search (SAMOS) method. The algorithm aims to find the best child network $\mathcal M$ by minimizing the error $p$ (penalty) of the child network. 

Initially, the SAMOS generates a feasible $\mathcal M$ as the starting point and gets the initial reward. In each round of the iteration, a neighbour of the child network of $\mathcal M$, denoted as $\mathcal M^k$, is generated. Then, based on the pretrained multi-path supernet, we can efficiently compute the reward of $\mathcal M^k$. Even if $\mathcal M^k$ does not have a better reward, it still has the opportunity to be accepted based on a probability of acceptance to avoid falling into a local minimum. With each iteration, the probability of acceptance decreases. After the loop, we obtain an approximately optimal child network and its corresponding reward. The details are shown in Algorithm~\ref{alg:SAMOS}.



\section{Experiment}
\subsection{Datasets}
The 300W-LP dataset~\cite{zhu2016face} is a large-pose database which consists of $60,000$ facial data synthesized by profiling method. 
To evaluate face alignment performance in the wild, we use three popular benchmarks, AFLW2000-3D ~\cite{zhu2016face}, AFLW-LFPA~\cite{jourabloo2016large} and Florence~\cite{bagdanov2011florence}.

The AFLW2000-3D contains 2,000 images from the AFLW dataset~\cite{Kstinger2011AnnotatedFL} and expands its annotations to 68 3D landmarks. In the evaluation set, there are 1,306, 462 and 232 samples for small, medium and large pose changes, respectively. 
The AFLW-LFPA dataset is another extension of the AFLW dataset, containing 1,299 face images with a balanced distribution of yaw angles and 34 landmark annotations. This database is evaluated on the task of sparse face alignment.
The Florence dataset contains 53 subjects with accurate 3D scans acquired from a structured-light scanning system. The evaluation focused on 3D dense face alignment.

\subsection{Implementation Details}
In the 3D face alignment tasks, we employ UV position maps as the regression targets. In supernet-based search space design, we set $L=10$ and $D=4$. The optimizer is Adam with a learning rate of 0.001. 

For SAMOS, the initial annealing temperature is set to  $2^{10}$ and the reduce rate is set to 0.85. The iteration times of the SAMOS is set to 200. We use the cplex optimizer to tune hyperparamters and obtain the optimal solution.

\subsection{Searched Network Architectures}
\begin{figure}[t]
\centering
\includegraphics[width=0.5\textwidth]{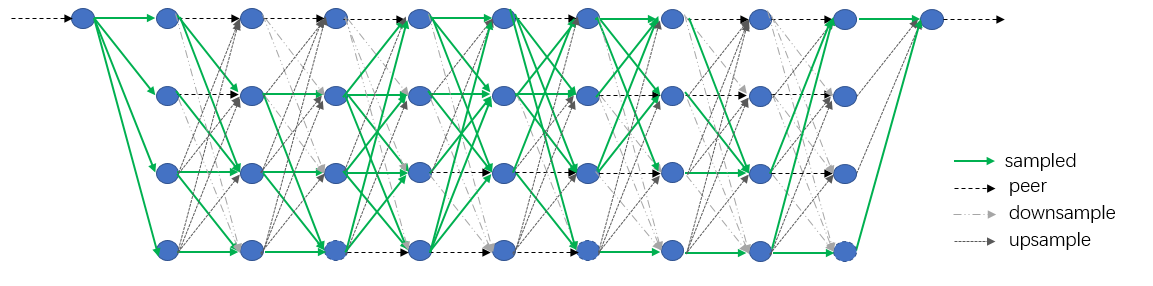}
\caption{Searched configuration for 3D face alignment.}
\label{fig:densesearchresults}
\end{figure}
One of the searched network architectures is presented in Fig.~\ref{fig:densesearchresults}. We roughly find that three different fusion progressive stages exist in the searched architecture, i.e, 
the downward-scale fusion stage, the full-scale fusion stage and the upward-scale fusion stage. 

The downward-scale stage primarily explores downward-direction connections of features, while the upward-scale stage tends to learn mostly upward-direction connections. The full-scale, however, involves intricate fusions of various scales. This phenomenon is intuitive since, in the downward-scale stage, relatively low-level features hold greater importance, and high-level semantic information is not well learned. Conversely, in the upward-scale stage, the high-level information becomes highly informative and dominant.

As manually designed multi-path networks such as HRNet \cite{sun2019deep}, Hourglass \cite{newell2016stacked} and PRNet \cite{feng2018joint} are special cases of the propose multi-path supernet, we also compare them with our searched network in Table~\ref{table:AFLW20003D-topo}. We find that the searched multi-path network significantly beats manually designed networks for face alignment. 

\begin{table}
\caption{Performance comparison on the AFLW2000-3D dataset for sparse face alignment using 2D coordinates. The NME{(\%)} with different yaw angles are reported.}
\label{table:AFLW20003D-topo}
\begin{center}
\resizebox{0.48\textwidth}{!}{
	\small
	\begin{tabular}{l|cccc|c}
		\hline
		Method &  [0, 30] & [30, 60] & [60, 90] & Mean NME &  NME \\
		\hline
		Hourglass~\cite{newell2016stacked} & 2.63 & 3.36 & 4.45 & 3.48 & --\\
		HRNet~\cite{sun2019deep} & 2.59 & 3.19 & 4.01 & 3.26 & -- \\
		PRNet~\cite{feng2018joint} & 2.75 & 3.51 & 4.61 & 3.62 & 3.26 \\
		\hline
		3DDFA~\cite{zhu2016face} & 3.78 & 4.54 & 7.93 & 5.42 & 6.03 \\
		DeFA~\cite{liu2017dense} & - & - & -& 4.50 & 4.36 \\
		3DSTN~\cite{bhagavatula2017faster} & 3.15 & 4.33 & 5.98 & 4.49 & --\\
		N-3DMM~\cite{tran2018nonlinear} & -& -& -& 4.12 & --\\
		SPDT~\cite{Piao2019SemiSupervisedM3} & 3.56 & 4.06 & 4.11 & 3.88 & --\\
		3DDFA V2~\cite{guo2020towards} & 2.63 & 3.42 & 4.48 & 3.51 & --\\
		SADRNet~\cite{Ruan2021SADRNetSD} & 2.66 & 3.30 & 4.42 & 3.46 & 3.05\\
		DSFNet-f~\cite{Li2023DSFNetDS} &  2.46  & 3.20  & 4.16 & 3.27 & -- \\
		\hline
		MONAS & \textbf{2.54} & \textbf{3.1} & \textbf{3.93} & \textbf{3.19} & \textbf{2.94} \\
		\hline
	\end{tabular}
}
\end{center}
\end{table}
\begin{table}[h!]
\caption{3D face alignment results on the AFLW-LFPA dataset. The NME is used for evaluation on visible landmarks.}
\label{table:AFLW_LFPA}
\begin{center}
\resizebox{0.46\textwidth}{!}{
	\small
	\begin{tabular}{c|c|c|c|c}
		\hline
		Method & DeFA~\cite{liu2017dense} & PRNet~\cite{feng2018joint} & HBCNN~\cite{Bulat2018HierarchicalBC}  & MONAS \\
		\hline
		NME & 3.86 & 2.93 & 3.02 & \textbf{2.71} \\
		\hline
	\end{tabular}
}
\end{center}
\end{table}
\subsection{Results of Sparse Alignment}
Following the settings of~\cite{feng2018joint}, we evaluate the performance of sparse face alignment on the point set of 68 landmarks, and compare our method with representative state-of-the-art methods on the AFLW2000-3D~\cite{zhu2016face} dataset. We follow~\cite{zhu2016face} to report the pose-specific alignment results. All methods are evaluated with Normalized Mean Error (NME). 

As shown in Table~\ref{table:AFLW20003D-topo}, our method outperforms recent state-of-the-art methods such as 3DDFA V2~\cite{guo2020towards}, PDT~\cite{Piao2019SemiSupervisedM3} and SADRNet~\cite{Ruan2021SADRNetSD} in all poses. Particularly, the proposed method decreases the error of 3DDFA V2~\cite{guo2020towards} by 0.09, 0.32 and 0.55 on faces in small, medium and large pose changes, respectively. The results demonstrate that the searched model dramatically improves the performance of large pose changes.

Results on the AFLW-LFPA dataset dataset are provided in Table~\ref{table:AFLW_LFPA}. Our approach also shows superior performance compared to existing state-of-the-art approaches.

\subsection{Results of Dense Alignment}
For dense 3D face alignment, we provide experimental results the Florence dataset~\cite{bagdanov2011florence}. Following the experimental settings in~\cite{jackson2017large,feng2018joint}, the face bounding boxes are calculated from the ground-truth point cloud. 
We choose the common face region with $19$K points to compare the performance. During evaluation, we first use Iterative Closest Points (ICP) algorithm to find the corresponding nearest points between the outputs of our method and ground truth, then calculate Mean Squared Error (MSE) normalized by the outer inter-ocular distance of 3D coordinates. The results are shown in Table~\ref{tab:nas_florence}.

We compare our approach with typical state-of-the-art methods such as 3DDFA V2~\cite{guo2020towards}, SPDT~\cite{Piao2019SemiSupervisedM3} and CMD~\cite{yuxiang2019}. Our method achieves the lowest NME of 3.51, surpassing existing approaches by clear margins. 

\begin{figure}
\centering
\includegraphics[width=0.35\textwidth]{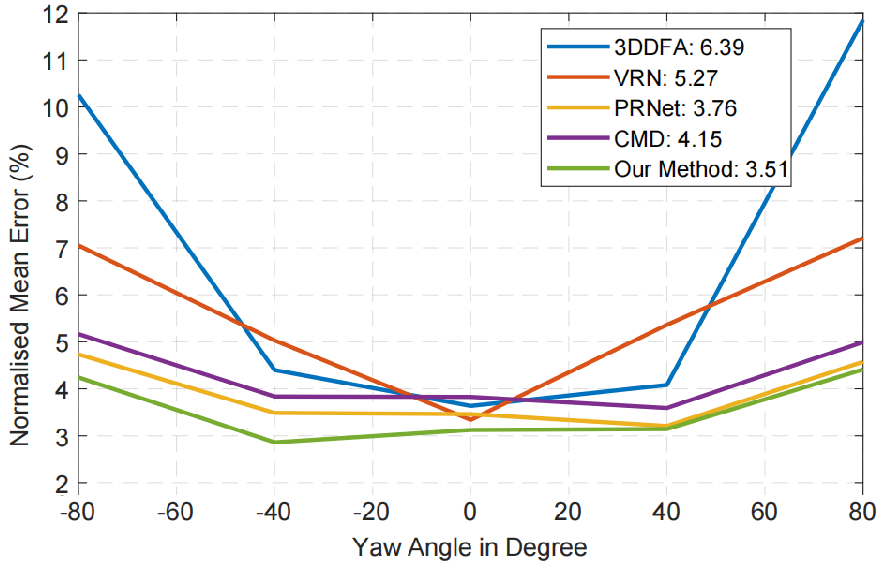}
\caption{Dense 3D face alignment results on the Florence dataset ($19$k vertices). The Normalized Mean Error of each method is showed in the legend.}
\label{fig:florence}
\end{figure}

To evaluate the reconstruction performance of our method across different poses, we calculate the NME under different yaw angles. As shown in Fig.~\ref{fig:florence}, most methods obtain good performance under the near frontal view. However, as the yaw angle increases, many methods fail to keep low error. Our approach obviously outperforms PRNet~\cite{feng2018joint} and CMD~\cite{yuxiang2019} under different profile views, and keeps relatively high performance under pose variations.

\begin{table}[t!]
\centering
\caption{3D dense alignment results on the Florence dataset.}
\label{tab:nas_florence}
\resizebox{0.3\textwidth}{!}{
\begin{tabular}{l|c|c}
	\toprule
	Method & Train data & NME \\
	\midrule
	3DDFA~\cite{zhu2016face} & 300WLP  & 6.39 \\
	VPN~\cite{jackson2017large} & 300WLP  & 5.27 \\
	PRNet~\cite{feng2018joint} & 300WLP  & 3.75 \\
	CMD~\cite{yuxiang2019} & 300WLP  & 4.15 \\
	SPDT~\cite{Piao2019SemiSupervisedM3} & 300WLP  & 3.83 \\
	3DDFA V2~\cite{guo2020towards} & 300WLP & 3.56 \\
	\hline
	MONAS & 300WLP & \textbf{3.51} \\
	\bottomrule
\end{tabular}}
\end{table}
\begin{table}[t!]
\caption{Ablation studies on the AFLW2000-3D dataset.}
\label{table:ablation}
\begin{center}
\resizebox{0.46\textwidth}{!}{
	\small
	\begin{tabular}{l|c|c|c}
		\hline
		Method & Training Cost & Search Cost & NME \\
		\hline
		Baseline  & 15 GPU days & 50 GPU hours &  3.88 \\ 
		w/o MNUS & 15 GPU days & 1.6 GPU hours &  3.69 \\ 
		w/o SAMOS & 3 GPU days & 50 GPU hours &  3.08 \\ 
		\hline
		MONAS  & 3 GPU days & 1.6 GPU hours & \textbf{2.94} \\
		\hline
	\end{tabular}
}
\end{center}
\end{table}
\begin{figure}
\centering
\includegraphics[width=0.46\textwidth]{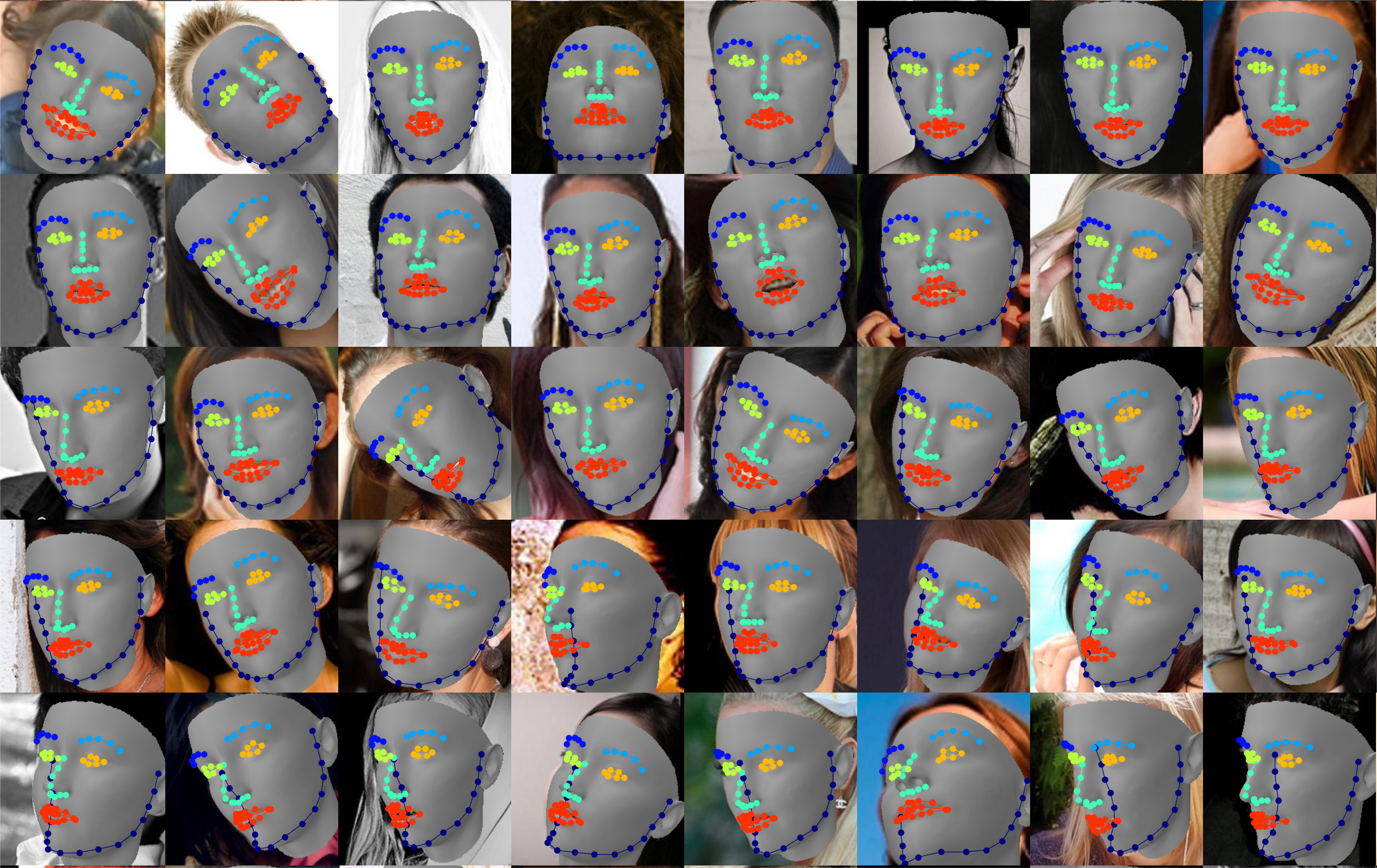}
\caption{Visualizations of 3D face alignment results with both sparse and dense landmarks.}
\label{fig:vis_result}
\end{figure}

\subsection{Ablations and Visualizations}
\noindent\textbf{Ablation Studies.} As the proposed MONAS mainly consists of two algorithms: the MNUS based supernet training and the SAMOS search algorithm, we provide ablation studies to demonstrate the effectiveness of both algorithms. The results on the AFLW2000-3D dataset is shown in Table~\ref{table:ablation}. The baseline method uses random sampling and random search, while w/o MNUS and w/o SAMOS denote the variants that only use the SAMOS based optimal structure search and the SAMOS based supernet training, respectively. The proposed MNUS and SAMOS significantly speed up the supernet training and the optimal network search, respectively. Both algorithms considerably improve the performance of face alignment. 

\noindent\textbf{Visualizations.} We present some visualizations of face alignment
results with various head poses on the AFLW-2000-3D dataset in Fig.~\ref{fig:vis_result}. The searched network gets robust results of both sparse and dense alignments under different poses. 
The proposed MONAS successfully detects the 3D facial keypoints and accurately constructs a 3D face model from a single profile image.

\section{Conclusions}
In this paper, we employ network architecture search (NAS) to tackle the sophisticated task of 3D face alignment and propose a novel Multi-path One-shot Neural Architecture Search (MONAS) framework to identify excelling multi-scale feature fusion models. To ensure the unbiasedness of possible feature connections during search, we introduce Multi-path Networks Unbiased Sampling (MNUS) strategy while generating child networks. To efficiently find the optimal network structure, two effective algorithms are designed to train the supernet and search the optimal network, respectively. Extensive experiments verify the effectiveness and efficiency of each component of the MONAS, and the proposed method achieves robust 3D face alignment under various head poses. We hope that this research can inspire solutions to face-related problems in computer vision, as well as advancements in network architecture search.

\bibliographystyle{IEEEtranS}
\bibliography{IEEEbib}
\end{document}